\documentclass[11pt]{article}
\usepackage[utf8]{inputenc}
\usepackage{coling2016}
\usepackage{times}
\usepackage{url}
\usepackage{latexsym}

\usepackage{amsmath}
\usepackage{booktabs}
\usepackage[inline]{enumitem}
\usepackage{lingmacros}
\usepackage{multirow}
\usepackage{graphicx}
\usepackage{subcaption}
\usepackage{tikz}
\usetikzlibrary{arrows,positioning}
\usepackage{pgfplots}
\pgfplotsset{compat=1.5}

\newcommand{\eps}{$\epsilon$}
\newcommand{\bos}{\_}%{$\langle$BEGIN$\rangle$}
\newcommand{\norma}{{Norma$^*$}}
\newcommand{\datareg}{\textsc{s}}
\newcommand{\dataaug}{\textsc{s+a}}

\title{Improving historical spelling normalization with bi-directional LSTMs and multi-task learning}
\author{Marcel Bollmann \\
  Department of Linguistics \\
  Ruhr-Universität Bochum \\
  Germany \\
  {\tt bollmann@linguistics.rub.de} \\\And
  Anders Søgaard \\
  Dpt.~of Computer Science \\
  University of Copenhagen \\
  Denmark \\
  {\tt soegaard@hum.ku.dk} \\}

\date{}

\begin{document}
\maketitle
\begin{abstract}
  Natural-language processing of historical documents is complicated by the
  abundance of variant spellings and lack of annotated data.  A common approach
  is to normalize the spelling of historical words to modern forms.  We explore
  the suitability of a deep neural network architecture for this task,
  particularly a deep bi-LSTM network applied on a character level.  Our model 
  compares well to previously established normalization
  algorithms when evaluated on a diverse set of texts from Early New High German.
  We show that multi-task learning with additional normalization data can 
  improve our model's performance further.
\end{abstract}

%
% The following footnote without marker is needed for the camera-ready
% version of the paper.
% Comment out the instructions (first text) and uncomment the 8 lines
% under "final paper" for your variant of English.
% 
\blfootnote{
    %
    % % final paper: en-uk version (to license, a licence)
    %
    % \hspace{-0.65cm}  % space normally used by the marker
    % This work is licensed under a Creative Commons 
    % Attribution 4.0 International Licence.
    % Licence details:
    % \url{http://creativecommons.org/licenses/by/4.0/}
    % 
    % % final paper: en-us version (to licence, a license)
    \hspace{-0.65cm}  % space normally used by the marker
    This work is licenced under a Creative Commons 
    Attribution 4.0 International License.
    License details:
    \url{http://creativecommons.org/licenses/by/4.0/}
}

\section{Introduction}
\label{sec:intro}

Interest in computational processing of historical documents is on the rise, as
evidenced by the growing field of digital humanities and the increasing number
of digitally available resources of historical data.  Spelling normalization,
i.e.\ the mapping of historical spelling variants to standardized/modernized
forms, is often employed as a pre-processing step to allow the utilization of
existing tools for the respective modern target language~\cite{Piotrowski2012}.

Training data for supervised learning of spelling normalization is typically
scarce in the historical domain.  Furthermore, dialectal influences and even individual preferences of an author can have a huge impact on the spelling characteristics in a particular text, meaning that even training data from other corpora of the same language and time period cannot always be reliably used.

Algorithms have often been developed with this fact in mind, e.g.\ by being based on some form of phonetic, graphematic, or semantic similarity measure~\cite{Jurish2010,Bollmann2012,Amoia2013}.  On the other hand, neural networks -- and
particularly deep networks with several hidden layers -- are assumed to work best when
trained on large amounts of data.  It is therefore not clear whether neural
networks are a good choice for this particular domain.

We frame spelling normalization as a character-based
sequence labeling task, and explore the suitability of a deep bi-directional long short-term memory model (bi-LSTM) in this setting.
By basing our model on individual characters as input, along with performing some basic preprocessing (e.g.,
downcasing all characters), we keep the vocabulary size small,
which in turn reduces the model's complexity and the amount of data required to
train it effectively.  We show that this model outperforms both the existing normalization tool Norma~\cite{Bollmann2012} and a CRF-based tagger when evaluated on a diverse dataset from Early New High German.

Furthermore, we experiment with a multi-task learning setup using auxiliary data that has similar, but not identical spelling characteristics to the target text.  We show that using bi-LSTMs with this multi-task learning setup can improve normalization accuracy further, while Norma and CRF do not profit much from the additional data in a traditional setup.

%%% Local Variables:
%%% mode: latex
%%% TeX-master: "master"
%%% End:

\section{Datasets}
\label{sec:data}

We use a total of 44~texts from the Anselm corpus~\cite{Dipper2013} of Early New High German.\footnote{\url{https://www.linguistics.rub.de/anselm/}}  The corpus is a collection of several manuscripts and prints of the same core text, a religious treatise.  Although the texts are semi-parallel and share some vocabulary, they were written in different time periods (between the 14th and 16th century) as well as different dialectal regions, and
show quite diverse spelling characteristics.  For example, the modern German word
\emph{Frau} `woman' can be spelled as \emph{fraw/vraw}~(Me), \emph{frawe}~(N2), \emph{frauwe}~(St), \emph{fraüwe}~(B2), \emph{frow}~(Stu), \emph{vrowe}~(Ka), \emph{vorwe}~(Sa), or \emph{vrouwe}~(B), among others.\footnote{Abbreviations in brackets refer to individual texts using the same internal IDs that are found in the Anselm corpus.}

All texts in the Anselm corpus are manually
annotated with gold-standard normalizations following guidelines described
in~\newcite{Krasselt2015}.  For our experiments, we excluded texts from the corpus that are shorter than 4,000~tokens, as well as a few texts for which annotations were not yet available at the time of writing (mostly Low German and Dutch versions).  Nonetheless, the remaining 44 texts are still quite short for machine-learning standards, ranging from about 4,200 to 13,200~tokens, with an average length of 7,353~tokens.

For all texts, we removed tokens that consisted solely of punctuation
characters.  We also lowercase all characters, since it helps keep the size of
the vocabulary low, and uppercasing of words is usually not very consistent in
historical texts.

\subsection{Conversion to labeled character sequences}
\label{sec:sequences}

Normalization is annotated on a word level; to reframe the problem as a character-based sequence labeling task, we need to align the historical wordforms and their normalizations on a character level.  Ideally, we would like these alignments to be linguistically plausible, i.e., characters that most likely correspond to each other (e.g., historical~\emph{j} and modern~\emph{i}, as in \emph{jn -- ihn} `him') should be aligned whenever possible.

The Levenshtein algorithm~\cite{Levenshtein1966} can be used to produce alignments that preferably align identical characters, but is ambiguous when multiple alignments with the same Levenshtein distance exist.  We therefore use iterated Levenshtein distance alignment~\cite{Wieling2009}, which uses pointwise mutual information on aligned segments to estimate statistical dependence, and favors alignments of characters that tend to cooccur often within the dataset.  Since different texts can use the same characters in different ways, we perform this iterated alignment separately for each text.

A difficulty of these alignments is that the two wordforms can be of different lengths.  We introduce a special epsilon label (\eps) whenever a historical character is not aligned to any character in the normalization.  We cannot do that for the inverse case, since the historical characters are our units of annotation and therefore need to be fixed, so we choose to perform a leftward merging of normalized characters whenever they are not aligned to any character in the historical wordform.  For the word-initial case, we introduce a special ``start of word'' symbol (\bos).  This symbol is prepended to each word during both training and testing, and is assigned the epsilon label during training when there is no word-initial insertion.

Here is an example of the final character sequence representation for the word pair \emph{vsfuret -- ausführt} `(he) leads out':

\enumsentence{\shortexnt{8}
{\bos & v & s & f & u  & r & e    & t}
{a    & u & s & f & üh & r & \eps & t }
}

A consequence of this approach is that our labels cannot only be characters, but also combinations of characters (such as \emph{üh} in the example above); our label set is therefore potentially unbounded.  However, we found that this is not much of a problem in practice, since these cases tend to be comparatively rare.

%%% Local Variables:
%%% mode: latex
%%% TeX-master: "master"
%%% End:

\section{Model}
\label{sec:model}

\begin{figure}[!t]
  \centering

\begin{tikzpicture}[scale=0.77,transform shape,>=stealth',auto]
  \tikzset{state/.style={draw,rectangle,minimum height=1.7em,minimum width=3.5em,
                         inner xsep=1em,inner ysep=0.5em,text depth=0.15em},
           emptystate/.style={inner sep=0.4em},
           outer/.style={outer sep=0},
           label/.style={align=center,font=\itshape\small}}

     \node[emptystate]      (I1) at (0, 0)         {{\bos}};
     \node[emptystate]      (I2) at (2, 0)         {{j}};
     \node[emptystate]      (I3) at (4, 0)         {{n}};

     \node[state]           (M1) at (0, 1.2)         {~};
     \node[state]           (M2) at (2, 1.2)         {~};
     \node[state]           (M3) at (4, 1.2)         {~};

     \node[emptystate]      (E0) at (-2, 2.7)        {~};
     \node[state]           (E1) at (0,  2.7)        {~};
     \node[state]           (E2) at (2,  2.7)        {~};
     \node[state]           (E3) at (4,  2.7)        {~};
     \node[emptystate]      (E4) at (6,  2.7)        {~};

     \node[emptystate]      (F0) at (-2, 3.9)        {~};
     \node[state]           (F1) at (0,  3.9)        {~};
     \node[state]           (F2) at (2,  3.9)        {~};
     \node[state]           (F3) at (4,  3.9)        {~};
     \node[emptystate]      (F4) at (6,  3.9)        {~};

     \node[emptystate]      (G0) at (-2, 5.1)        {~};
     \node[state]           (G1) at (0,  5.1)        {~};
     \node[state]           (G2) at (2,  5.1)        {~};
     \node[state]           (G3) at (4,  5.1)        {~};
     \node[emptystate]      (G4) at (6,  5.1)        {~};

     \node[state]           (P1) at (0, 6.6)         {~};
     \node[state]           (P2) at (2, 6.6)         {~};
     \node[state]           (P3) at (4, 6.6)         {~};

	 \node[emptystate]      (O1) at (0, 7.8)         {\eps};
     \node[emptystate]      (O2) at (2, 7.8)         {ih};
     \node[emptystate]      (O3) at (4, 7.8)         {n};

     \node[emptystate]      (L1) at (-4.0, 1.2)      {\emph{embedding layer}};
     \node[emptystate]      (L2) at (-4.0, 4.1)      {\emph{stack of}};
     \node[emptystate]      (L3) at (-4.0, 3.5)      {\emph{bi-LSTM layers}};
     \node[emptystate]      (L4) at (-4.0, 6.6)      {\emph{prediction layer}};

     \draw [->]  (I1) to (M1);
     \draw [->]  (I2) to (M2);
     \draw [->]  (I3) to (M3);
	 \draw [->]  (M1) to (E1);
	 \draw [->]  (M2) to (E2);
	 \draw [->]  (M3) to (E3);

	 \coordinate[above=0.15 of E4.west]  (E4_tw);
	 \coordinate[above=0.15 of E3.east]  (E3_te);
	 \coordinate[above=0.15 of E3.west]  (E3_tw);
	 \coordinate[above=0.15 of E2.east]  (E2_te);
	 \coordinate[above=0.15 of E2.west]  (E2_tw);
	 \coordinate[above=0.15 of E1.east]  (E1_te);
     \coordinate[below=0.15 of E3.west]  (E3_bw);
     \coordinate[below=0.15 of E2.east]  (E2_be);
     \coordinate[below=0.15 of E2.west]  (E2_bw);
     \coordinate[below=0.15 of E1.east]  (E1_be);
     \coordinate[below=0.15 of E1.west]  (E1_bw);
     \coordinate[below=0.15 of E0.east]  (E0_be);

	 \coordinate[above=0.15 of F4.west]  (F4_tw);
	 \coordinate[above=0.15 of F3.east]  (F3_te);
	 \coordinate[above=0.15 of F3.west]  (F3_tw);
	 \coordinate[above=0.15 of F2.east]  (F2_te);
	 \coordinate[above=0.15 of F2.west]  (F2_tw);
	 \coordinate[above=0.15 of F1.east]  (F1_te);
     \coordinate[below=0.15 of F3.west]  (F3_bw);
     \coordinate[below=0.15 of F2.east]  (F2_be);
     \coordinate[below=0.15 of F2.west]  (F2_bw);
     \coordinate[below=0.15 of F1.east]  (F1_be);
     \coordinate[below=0.15 of F1.west]  (F1_bw);
     \coordinate[below=0.15 of F0.east]  (F0_be);

	 \coordinate[above=0.15 of G4.west]  (G4_tw);
	 \coordinate[above=0.15 of G3.east]  (G3_te);
	 \coordinate[above=0.15 of G3.west]  (G3_tw);
	 \coordinate[above=0.15 of G2.east]  (G2_te);
	 \coordinate[above=0.15 of G2.west]  (G2_tw);
	 \coordinate[above=0.15 of G1.east]  (G1_te);
     \coordinate[below=0.15 of G3.west]  (G3_bw);
     \coordinate[below=0.15 of G2.east]  (G2_be);
     \coordinate[below=0.15 of G2.west]  (G2_bw);
     \coordinate[below=0.15 of G1.east]  (G1_be);
     \coordinate[below=0.15 of G1.west]  (G1_bw);
     \coordinate[below=0.15 of G0.east]  (G0_be);

	 \draw [densely dotted,->]  (E4_tw) to (E3_te);
     \draw [densely dotted,->]  (E3_tw) to (E2_te);
     \draw [densely dotted,->]  (E2_tw) to (E1_te);
	 \draw [densely dotted,->]  (E0_be) to (E1_bw);
     \draw [densely dotted,->]  (E1_be) to (E2_bw);
     \draw [densely dotted,->]  (E2_be) to (E3_bw);

	 \draw [->] (E1) to (F1);
	 \draw [->] (E2) to (F2);
	 \draw [->] (E3) to (F3);
     
	 \draw [densely dotted,->]  (F4_tw) to (F3_te);
     \draw [densely dotted,->]  (F3_tw) to (F2_te);
     \draw [densely dotted,->]  (F2_tw) to (F1_te);
	 \draw [densely dotted,->]  (F0_be) to (F1_bw);
     \draw [densely dotted,->]  (F1_be) to (F2_bw);
     \draw [densely dotted,->]  (F2_be) to (F3_bw);

	 \draw [->] (F1) to (G1);
	 \draw [->] (F2) to (G2);
	 \draw [->] (F3) to (G3);
     
	 \draw [densely dotted,->]  (G4_tw) to (G3_te);
     \draw [densely dotted,->]  (G3_tw) to (G2_te);
     \draw [densely dotted,->]  (G2_tw) to (G1_te);
	 \draw [densely dotted,->]  (G0_be) to (G1_bw);
     \draw [densely dotted,->]  (G1_be) to (G2_bw);
     \draw [densely dotted,->]  (G2_be) to (G3_bw);

	 \draw [->]  (G1) to (P1);
	 \draw [->]  (G2) to (P2);
	 \draw [->]  (G3) to (P3);
	 \draw [->]  (P1) to (O1);
	 \draw [->]  (P2) to (O2);
	 \draw [->]  (P3) to (O3);

\end{tikzpicture}

\caption{Flow diagram of the bi-LSTM character sequence labeling model, unrolled for time, for the word pair \emph{jn} -- \emph{ihn} `him'.}
  \label{fig:model}
\end{figure}
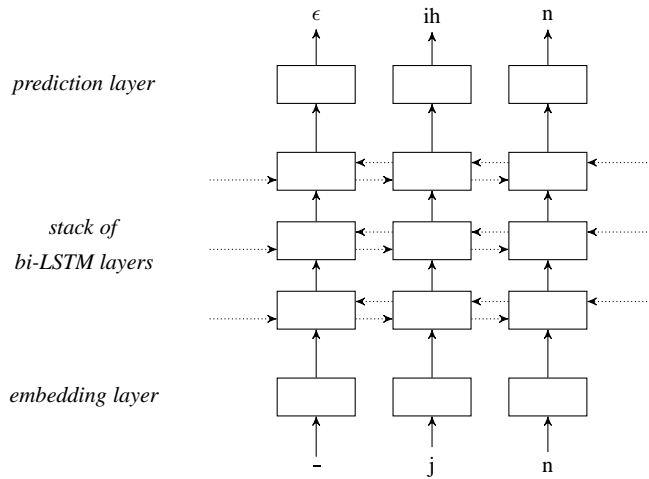

Our model architecture consists of:
\begin{enumerate*}[label=(\roman*)]
  \item an embedding layer for the input characters;
  \item a stack of bi-directional long short-term memory units (bi-LSTMs); and
  \item a final dense layer with a softmax activation to generate a probability distribution over the output classes at each timestep.
\end{enumerate*}
An illustration of the model can be found in Figure~\ref{fig:model}.

The embedding layer maps one-hot input vectors (representing historical characters) to dense vectors.  We did not use pre-trained embeddings; the embeddings are initialized randomly and learned as part of the regular network training process.

LSTMs \cite{Hochreiter1997} are a form of recurrent neural
network~(RNN) designed to better learn long-term dependencies, and have proven advantageous to
plain RNNs on many tasks.  Bi-directional LSTMs read their input in both normal and reversed order, allowing the model to learn from both left and right context at each input timestep.  A stack of bi-LSTMs, or a deep bi-LSTM, is a configuration of several bi-LSTM units so that the output of the $i$th unit is the input of the $(i+1)$th unit.  In our model, we use a stack of three bi-LSTM layers.

%%% Local Variables:
%%% mode: latex
%%% TeX-master: "master"
%%% End:

The final dense layer is used to generate the output predictions, based on a linear transformation of the bi-LSTM outputs for each timestep followed by a softmax activation.  We train the model by minimizing the cross-entropy loss across all output characters, and using backpropagation to update the weights in all layers (including the embedding layer).  During prediction, we generate output labels in a greedy fashion, choosing the label with the highest probability for each timestep.

\subsection{Multi-task learning setup}

In multi-task learning~(MTL), the performance of a model on a given task is improved by additionally training it on one or more auxiliary tasks~\cite{Caruana1993}.  For our bi-LSTM model, this means that all layers of the model are shared between the tasks apart from the final prediction layer, which is kept separate for the main and auxiliary tasks.  This way, errors in an auxiliary task that are backpropagated through the network also affect the prediction of the main task, helping to regularize the network's weights and prevent overfitting.  

Multi-task learning with (deep) neural network architectures was shown to be effective for a variety of NLP tasks, such as part-of-speech tagging, chunking, named entity recognition~\cite{Collobert2011}; sentence compression~\cite{Klerke2016}; or machine translation~\cite{Luong2016}.

In our experiments, we regard spelling normalization within a target domain (i.e., a given historical text) as our main task, while using normalization within related domains (i.e., texts from a similar time period, but with distinct spelling characteristics) as our auxiliary task.  During training, we alternate between training on a random instance from the main and the auxiliary tasks.

\subsection{Hyperparameters}
\label{sec:model-hyper}

We set aside one of the texts (B) from the Anselm corpus for testing different hyperparameter configurations.  On this text, we achieved the best results with a dimensionality of~128 for the embedding and bi-LSTM layers, using a dropout of~0.1, and training the model for 30~iterations.  These settings were subsequently used for all further experiments.

\subsection{Other models used for comparison}

For comparison, we also train and evaluate with the Norma tool described by~\newcite{Bollmann2012},
since it was originally developed for the Anselm corpus and the
implementation is publicly
available.\footnote{\url{https://github.com/comphist/norma}} However, Norma actually
consists of a combination of three different normalization methods, one of which is a simple wordlist mapping of historical tokens to normalized forms.  Since this wordlist mapping is conceptually very simple and could easily be added to our (or any other) normalization method, we exclude it for the comparison, and only use Norma's remaining two algorithms (which we denote \norma).

Additionally, since we frame the problem as a sequence labeling task, we compare our results to a simple sequence labeling model using conditional random fields~(CRF). The CRF model gets the same input/output sequences as our bi-LSTM model (cf.\ Sec.~\ref{sec:sequences}), and uses the two preceding and following characters from the historical wordform as additional features.  Implementation was done with CRFsuite~\cite{CRFsuite} using the averaged perceptron algorithm for training.

%%% Local Variables:
%%% mode: latex
%%% TeX-master: "master"
%%% End:

\section{Evaluation}
\label{sec:eval}

\begin{table}
\centering\small
\begin{tabular}{llrrrrrrr}
\toprule
\textbf{ID} & \textbf{Region} & \textbf{Tokens} & \multicolumn{2}{c}{\textbf{\norma}} & \multicolumn{2}{c}{\textbf{CRF}} & \multicolumn{2}{c}{\textbf{Bi-LSTM}} \\
\cmidrule(lr){4-5}\cmidrule(lr){6-7}\cmidrule(lr){8-9}
& & & \multicolumn{1}{c}{\datareg} & \multicolumn{1}{c}{\dataaug} & \multicolumn{1}{c}{\datareg} & \multicolumn{1}{c}{\dataaug} & \multicolumn{1}{c}{\datareg} & \multicolumn{1}{c}{\dataaug$^\dagger$} \\
\midrule
B & East Central & 4,718 & 80.30\% & 77.80\% & 76.30\% & 72.80\% & 79.20\% & \textbf{81.70\%}\\
D3 & East Central & 5,704 & 80.50\% & 80.20\% & 77.20\% & 73.00\% & 80.10\% & \textbf{81.20\%}\\
H & East Central & 8,427 & 82.70\% & 82.90\% & 78.60\% & 76.00\% & \textbf{85.00\%} & 82.30\%\\
\midrule
B2 & West Central & 9,145 & 76.10\% & 77.60\% & 74.60\% & 71.70\% & \textbf{82.00\%} & 79.60\%\\
KÄ1492 & West Central & 7,332 & 77.50\% & 74.40\% & 74.80\% & 68.40\% & \textbf{81.60\%} & 80.50\%\\
KJ1499 & West Central & 7,330 & 77.00\% & 72.90\% & 73.50\% & 68.40\% & \textbf{84.50\%} & 79.20\%\\
N1500 & West Central & 7,272 & 76.70\% & 75.30\% & 72.70\% & 67.20\% & 79.00\% & \textbf{79.20\%}\\
N1509 & West Central & 7,418 & 78.10\% & 73.30\% & 74.30\% & 68.80\% & \textbf{80.80\%} & 80.10\%\\
N1514 & West Central & 7,412 & 78.30\% & 73.80\% & 72.20\% & 69.90\% & 79.00\% & \textbf{80.10\%}\\
St & West Central & 7,407 & 72.60\% & 73.80\% & 70.30\% & 68.70\% & \textbf{75.50\%} & 75.20\%\\
\midrule
D4 & Upper/Central & 5,806 & 75.60\% & 75.60\% & 72.40\% & 70.90\% & 76.50\% & \textbf{76.60\%}\\
N4 & Upper & 8,593 & 78.20\% & 78.10\% & 80.00\% & 78.40\% & 81.80\% & \textbf{83.40\%}\\
s1496/97 & Upper & 5,840 & 81.70\% & 83.40\% & 77.70\% & 76.90\% & 83.00\% & \textbf{84.10\%}\\
\midrule
B3 & East Upper & 6,222 & 80.80\% & 80.60\% & 79.50\% & 79.10\% & 81.50\% & \textbf{83.20\%}\\
Hk & East Upper & 8,690 & 77.80\% & 79.30\% & 78.20\% & 77.90\% & 80.90\% & \textbf{82.20\%}\\
M & East Upper & 8,700 & 74.30\% & 74.40\% & 72.80\% & 68.40\% & \textbf{83.90\%} & 80.90\%\\
M2 & East Upper & 8,729 & 75.80\% & 76.00\% & 75.10\% & 72.40\% & 76.70\% & \textbf{80.20\%}\\
M3 & East Upper & 7,929 & 79.00\% & 79.70\% & 77.30\% & 74.10\% & \textbf{80.40\%} & 79.60\%\\
M5 & East Upper & 4,705 & 80.60\% & 80.70\% & 76.40\% & 78.30\% & 77.70\% & \textbf{82.90\%}\\
M6 & East Upper & 4,632 & 75.90\% & 76.30\% & 73.70\% & 74.40\% & 75.20\% & \textbf{79.30\%}\\
M9 & East Upper & 4,739 & 82.20\% & 81.50\% & 79.00\% & 76.90\% & 80.40\% & \textbf{83.60\%}\\
M10 & East Upper & 4,379 & 77.00\% & 78.60\% & 76.00\% & 75.80\% & 75.10\% & \textbf{81.30\%}\\
Me & East Upper & 4,560 & 79.70\% & 80.10\% & 76.90\% & 75.50\% & 80.30\% & \textbf{83.70\%}\\
Sb & East Upper & 7,218 & 78.00\% & 76.60\% & 75.70\% & 74.80\% & \textbf{80.00\%} & 78.50\%\\
T & East Upper & 8,678 & 76.70\% & 78.60\% & 73.40\% & 72.20\% & 75.80\% & \textbf{79.00\%}\\
W & East Upper & 8,217 & 75.90\% & 78.30\% & 78.20\% & 77.00\% & \textbf{81.40\%} & 80.80\%\\
We & East Upper & 6,661 & \textbf{83.10\%} & 81.50\% & 78.60\% & 75.80\% & 81.50\% & \textbf{83.10\%}\\
\midrule
Ba & North Upper & 5,934 & 79.80\% & 81.20\% & 80.20\% & 78.70\% & 80.70\% & \textbf{82.80\%}\\
Ba2 & North Upper & 5,953 & 81.40\% & 80.00\% & 78.10\% & 77.90\% & 82.50\% & \textbf{84.10\%}\\
M4 & North Upper & 8,574 & 76.90\% & 76.70\% & 75.70\% & 75.00\% & 79.40\% & \textbf{82.30\%}\\
M7 & North Upper & 4,638 & 79.40\% & 79.80\% & 75.60\% & 74.20\% & 78.20\% & \textbf{82.10\%}\\
M8 & North Upper & 8,275 & 78.50\% & 77.00\% & 78.20\% & 78.40\% & 81.10\% & \textbf{82.50\%}\\
n & North Upper & 9,191 & 79.60\% & 81.30\% & 81.90\% & 78.20\% & 84.40\% & \textbf{84.70\%}\\
N & North Upper & 13,285 & 75.50\% & 76.30\% & 71.70\% & 68.90\% & \textbf{79.00\%} & 76.90\%\\
N2 & North Upper & 7,058 & 82.20\% & 81.90\% & 80.30\% & 81.60\% & \textbf{84.30\%} & 83.40\%\\
N3 & North Upper & 4,192 & 79.10\% & 80.80\% & 76.40\% & 77.50\% & 77.60\% & \textbf{84.20\%}\\
\midrule
Be & West Upper & 8,203 & 75.50\% & 76.40\% & 75.30\% & 73.40\% & \textbf{78.80\%} & 78.00\%\\
Ka & West Upper & 12,641 & 73.80\% & 74.10\% & 75.40\% & 72.80\% & 80.10\% & \textbf{80.30\%}\\
SG & West Upper & 7,838 & 80.10\% & 79.90\% & 78.00\% & 76.80\% & \textbf{81.70\%} & 80.90\%\\
Sa & West Upper & 8,668 & 72.60\% & 73.50\% & 71.90\% & 71.40\% & 76.10\% & \textbf{76.50\%}\\
St2 & West Upper & 8,834 & 73.20\% & 73.40\% & 73.20\% & 73.00\% & 78.20\% & \textbf{79.90\%}\\
Stu & West Upper & 8,686 & 77.70\% & 77.10\% & 76.50\% & 72.10\% & \textbf{79.40\%} & 77.00\%\\
Sa2 & West Upper & 8,011 & 77.50\% & 77.90\% & 73.50\% & 73.30\% & 79.50\% & \textbf{79.70\%}\\
\midrule
Le & Dutch & 7,087 & 69.50\% & 60.30\% & 65.00\% & 55.80\% & \textbf{75.60\%} & 67.50\%\\
\midrule
\textit{Average} & & 7,353 & 77.83\% & 77.48\% & 75.73\% & 73.70\% & 79.90\% & \textbf{80.55\%} \\
\bottomrule
\end{tabular}
  \caption{Word accuracy on the Anselm dataset, evaluated on the first 1,000~tokens; \datareg{}~= training set from the same text, \dataaug{}~= like \datareg{}, but augmented with 10,000~tokens randomly sampled from the other texts; $^\dagger$~=~Bi-LSTM (\dataaug{}) is the multi-task learning setup.  Best results shown in bold.}
  \label{tab:results}
\end{table}

%%% Local Variables:
%%% mode: latex
%%% TeX-master: "master"
%%% End:

We evaluate our model separately for each text in our dataset.  From each text, we use 1,000~tokens as our evaluation set, set aside another 1,000~tokens as a development set (which was not currently used), and train on the remaining tokens (between 2,000 and 11,000, depending on the text).  Both CRF and our bi-LSTM model get their input as character sequences (as described in Sec.~\ref{sec:sequences}), while Norma requires full words as input.

For the multi-task learning setup, we randomly sample from all Anselm texts and regard each text as its own task.  Effectively, we are learning a joint model over all Anselm texts with shared parameters but distinct prediction layers, while viewing the text we are currently evaluating on as our main task and the others as auxiliary tasks.  The MTL~setup is only applicable to our bi-LSTM model; however, since the auxiliary task consists of spelling normalization with data from the same corpus (although with a higher variety of different spelling characteristics compared to the target text), it is possible that the other methods could also profit from this additional training data.  We therefore also evaluate Norma and CRF when the training sets have been augmented by 10,000~randomly sampled training examples from all texts.

\subsection{Word accuracy}
\label{sec:eval-acc}

Evaluation results in terms of word-level accuracy are presented in
Table~\ref{tab:results}.

Columns ``\datareg'' show results for the traditional setup without multi-task learning.  The basic bi-LSTM model performs better than Norma on 34 of the 44~texts.  On average, there is an increase of $2.1$ percentage points~(pp), although the differences on individual texts vary wildly, from $-2.9$ pp (M5) to $+9.6$ pp (M), giving a standard deviation of $2.7$ pp.  The CRF model, on the other hand, is almost always worse than Norma, averaging a difference of $-2.1$ pp ($\pm 2.0$).  This indicates that the reformulation of the task as character-based sequence labeling cannot alone be responsible for the bi-LSTM results, but the choice of a neural network architecture is crucial, too.

Columns ``\dataaug'' present the results when using the augmented training set.  For bi-LSTM, this is the multi-task learning setup---using MTL improves the results by $+0.7$~pp ($\pm 2.8$) on average, but again there is a high variance within the individual scores.  However, for the other methods, adding the 10,000~randomly selected samples to the training set actually decreases the average accuracy, by $-0.4$~pp for Norma and $-2.0$~pp for CRF.  This is likely due to the fact that this additional training set introduces a variety of spelling characteristics that are not found in the target text.  While Norma and CRF cannot handle this out-of-domain training data well, the MTL~setup can actually profit from it in many cases.

Table~\ref{tab:results} also shows a rough classification of the dialectal regions from which the texts originate.  There is a slight trend for multi-task learning to be advantageous on texts from the East and North Upper German regions, while for the Central and West Upper German texts, there are more instances of the standard bi-LSTM model (\datareg) being better than the MTL~model (\dataaug).  This could either be due to linguistic properties of these dialectal regions, or due to the fact that East/North Upper German texts make up the majority of the dataset, thereby also featuring more prominently in the ``\dataaug'' settings.

The latter hypothesis is supported by the case of the `Le'~text, which is the only Dutch text in the sample (but which was nonetheless normalized to modern German in the corpus).  Here, the ``\dataaug''~settings of the experiments all show a dramatic decrease in accuracy (up to $-9.2$~pp), suggesting that it is disadvantageous to augment the training set with samples that are too different from the target domain, even for the MTL~setup.

In general, however, one of the bi-LSTM models is always best; there is only one text (We) for which Norma achieves an equal accuracy.  This indicates that deep neural networks can be applied successfully to the spelling normalization task even with a comparatively small amount of training data.  Also, we note that Norma always requires a lexical resource which it uses to filter results, while we do not.

\subsection{Effect of training set size}
\label{sec:eval-size}

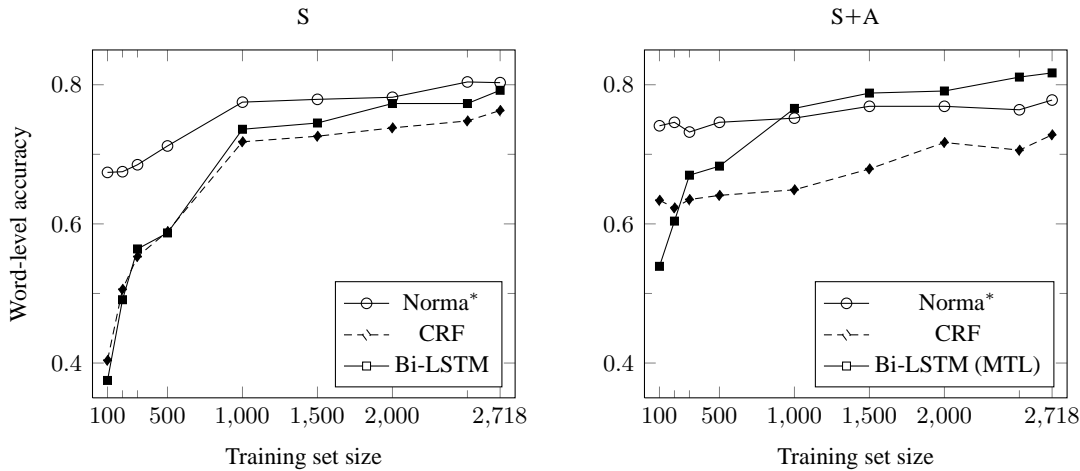
\begin{figure}\centering
\pgfplotsset{
tick label style={font=\small},
label style={font=\small},
legend style={font=\footnotesize}
}
%\pgfplotsset{every axis/.append style={line width=1pt}}

  \begin{subfigure}[t]{0.45\textwidth}
    %\centering
	\begin{tikzpicture}
	\begin{axis}[
        xmin=0,
        xmax=2818,
        xtick={100,500,1000,1500,2000,2718},
        minor xtick={200,300,2500},
        minor y tick num=1,
        ymin=0.35,
        ymax=0.85,
		xlabel={Training set size},
		ylabel={Word-level accuracy},
        title={\datareg},
        legend pos=south east,
        width=1.0\textwidth
    ]
	\addplot[mark=o,style=solid] coordinates {  % Norma
		(100, 0.674)
        (200, 0.675)
        (300, 0.685)
        (500, 0.712)
        (1000, 0.775)
        (1500, 0.779)
        (2000, 0.782)
        (2500, 0.804)
        (2718, 0.803)
	};
    \addlegendentry{\norma}
	\addplot[mark=diamond*,mark size=2pt,style=densely dashed] coordinates {  % CRF
		(100, 0.404)
        (200, 0.506)
        (300, 0.553)
        (500, 0.589)
        (1000, 0.718)
        (1500, 0.726)
        (2000, 0.738)
        (2500, 0.748)
        (2718, 0.763)
	};
    \addlegendentry{CRF}
	\addplot[mark=square*,mark size=1.5pt,style=solid] coordinates {  % LSTM
		(100, 0.375)
        (200, 0.491)
        (300, 0.564)
        (500, 0.587)
        (1000, 0.736)
        (1500, 0.745)
        (2000, 0.773)
        (2500, 0.773)
        (2718, 0.792)
	};
    \addlegendentry{Bi-LSTM}
	\end{axis}
\end{tikzpicture}
    %\caption{Regular training set (\datareg)}
    \label{fig:sizes-reg}
  \end{subfigure}
\quad
  \begin{subfigure}[t]{0.45\textwidth}
    \centering
	\begin{tikzpicture}
	\begin{axis}[
        xmin=0,
        xmax=2818,
        xtick={100,500,1000,1500,2000,2718},
        minor xtick={200,300,2500},
        minor y tick num=1,
        ymin=0.35,
        ymax=0.85,
		xlabel={Training set size},
        title={\dataaug},
        legend pos=south east,
        width=1.0\textwidth
    ]
	\addplot[mark=o,style=solid] coordinates {  % Norma
		(100, 0.741)
        (200, 0.746)
        (300, 0.732)
        (500, 0.746)
        (1000, 0.752)
        (1500, 0.769)
        (2000, 0.769)
        (2500, 0.764)
        (2718, 0.778)
	};
    \addlegendentry{\norma}
	\addplot[mark=diamond*,mark size=2pt,style=densely dashed] coordinates {  % CRF
		(100, 0.634)
        (200, 0.623)
        (300, 0.635)
        (500, 0.641)
        (1000, 0.649)
        (1500, 0.679)
        (2000, 0.717)
        (2500, 0.706)
        (2718, 0.728)
	};
    \addlegendentry{CRF}
	\addplot[mark=square*,mark size=1.5pt,style=solid] coordinates {  % LSTM
		(100, 0.539)
        (200, 0.604)
        (300, 0.670)
        (500, 0.683)
        (1000, 0.766)
        (1500, 0.788)
        (2000, 0.791)
        (2500, 0.811)
        (2718, 0.817)
	};
    \addlegendentry{Bi-LSTM (MTL)}
	\end{axis}
\end{tikzpicture}
    %\caption{Augmented training set/multi-task learning (\dataaug)}
    \label{fig:sizes-aug}
  \end{subfigure}

  \caption{Word accuracy on the `B'~text for different sizes of the training set; left~= train only on the training set from `B' (\datareg); right~= use augmented training set/multi-task learning (\dataaug).}
\label{fig:sizes}
\end{figure}

In our evaluation, we use all but the first 2,000~tokens from a text for training (cf.\ the beginning of Sec.~\ref{sec:eval}).  Consequently, the training sets for each text are of different sizes.  We calculate Spearman's rank correlation coefficient ($\rho$) between the size of the training sets and the normalization accuracy for each column in Table~\ref{tab:results}.  We find no significant correlation for the CRF and bi-LSTM models ($|\rho| < 0.25$), although there seems to be a moderate \emph{inverse} correlation for the Norma results ($\rho \approx -0.48$ on Norma ``\datareg'').  The reasons for this are beyond the scope of this paper, though.

The question of how much training data is needed to effectively train a model is particularly relevant for historical spelling normalization, since training data can be very sparse in this domain.  We therefore choose to evaluate each method in a scenario where we consider a single text, but vary the size of the training set, to estimate how well they perform with fewer data.

Figure~\ref{fig:sizes} shows the results for different training set sizes on the `B'~text.  Not surprisingly, when training on only 100~tokens, accuracy is bad ($<41\%$) for CRF and bi-LSTM.  Norma, on the other hand, already achieves $67.4\%$ in this scenario.  The biggest gains for all three methods can be seen for training set sizes between 100 and 1,000~tokens---for larger set sizes, the gains become less, and all three methods are within close range of each other.

For the ``\dataaug'' scenario, all models have noticeably higher accuracy even with only 100~tokens from the `B'~text.  However, the increases for Norma and CRF are not as high as in the ``\datareg''~scenario; this is not surprising, since the total training set for these methods always contains at least 10,000~tokens (from the auxiliary set), and it is only the proportion of tokens coming from the `B'~text that increases.  The bi-LSTM model with multi-task learning behaves differently, though: while it starts off as the weakest model (on 100~tokens), it is the best model when training on 1,000~tokens or more.

These learning curves illustrate that the MTL~setup is fundamentally different from adding the auxiliary data to the training set normally, as is the case with CRF and the Norma tool.  They also show that our bi-LSTM models can be better than or at least competitive with CRF/Norma for training set sizes as low as 1,000~tokens.

\subsection{Multi-task learning with grapheme-to-phoneme mappings}
\label{sec:celex}

It is conceivable to use different tasks than historical spelling normalization as the auxiliary task in a multi-task learning setup.  In particular, we also experimented with grapheme-to-phoneme mapping as the auxiliary task, since it can be seen as a similar form of character-based sequence transduction.

For our dataset, we used the German part of the CELEX lexical data\-base~\cite{CELEX}, particularly the database of phonetic transcriptions of German wordforms.  The database contains a total of 365,530~wordforms with transcriptions in DISC~format, which assigns one character to each distinct phonological segment (including affricates and diphthongs).  For example, the word \emph{Jungfrau} `virgin' is represented as \texttt{'jUN-frB}.  We randomly sampled 4,000~tokens from this dataset for our experiment, and used the same algorithm as for the historical data to convert these mappings to a character-based sequence representation (cf.\ Sec.~\ref{sec:sequences}).

The evaluation, however, showed no real benefit of this MTL~setup compared to the bi-LSTM model without MTL.  
While accuracy increased for some texts by up to $2.6$~pp, it decreased slightly for the majority of texts, averaging to a $-0.4$~pp difference to the basic model.

%%% Local Variables:
%%% mode: latex
%%% TeX-master: "master"
%%% End:

\section{Related Work}
\label{sec:related}

Various methods have been proposed to perform spelling normalization on
historical texts; for an overview, see \newcite{Piotrowski2012}.  Many approaches use edit distance calculations or some form
of character-level rewrite rules, but require either hand-crafting of the
rules~\cite{Baron2008} or a lexical resource to filter their
output~\cite{Bollmann2012,Porta2013}.

A newer approach is the application of character-based statistical machine
translation~\cite{Pettersson2013,Sanchez-Martinez2013,Scherrer2013}.  In contrast to our sequence labeling approach, these methods do not require a fixed character alignment between wordforms, but it is not clear whether this is actually an advantage.  To our knowledge, a comparative evaluation between these methods and other approaches has not yet been done.

\newcite{AlAzawi2013} present the only other approach we are aware of that
applies neural networks to normalization of historical data.  They also use bi-directional LSTMs, but differ from our approach in the way they perform alignment between historical and modern wordforms.  More importantly, they evaluate their model on a single dataset, the Luther bible, which has much more regular spelling than the texts in the Anselm corpus and is also significantly longer: they use about 200,000~tokens for their training set.

%%% Local Variables:
%%% mode: latex
%%% TeX-master: "master"
%%% End:

\section{Conclusion and Future Work}
\label{sec:end}

We presented an approach to historical spelling normalization using bi-directional long short-term memory networks and showed that it outperforms a CRF baseline and the Norma tool
by~\newcite{Bollmann2012} for almost all of the texts in our dataset, a diverse corpus of Early New High German, despite using a relatively low amount of training data (about 2,000 to 11,000~tokens) and not making use of a lexical resource (like Norma does).  We showed further that multi-task learning with additional normalization data can improve accuracy with bi-LSTMs, while adding the same data to the training set of Norma and CRF does not help on average, and can even be detrimental.

Many improvements to this approach are conceivable.  Character-based statistical machine translation has been successfully applied to spelling normalization (cf.\ Sec.~\ref{sec:related}), but we are not aware of any experiments with neural machine translation~\cite{Cho2014} on this domain.  Using an encoder--decoder architecture, e.g.\ similar to \newcite{Sutskever2014}, would remove the need for an explicit character alignment (cf.\ Sec.~\ref{sec:sequences}) but could also make the model more complex and potentially more difficult to train, so it is unclear whether this would be an improvement to our approach.

With regard to multi-task learning, our results seem to indicate that for the auxiliary task, it is preferable to use data with similar characteristics to the data in the main task.  On the other hand, depending on the language variety to be annotated, such data might not always be readily available.  We would therefore like to do further experiments with auxiliary data from different corpora or even different string transduction tasks, to see if and under which conditions they can have a beneficial effect on the spelling normalization task.

%%% Local Variables:
%%% mode: latex
%%% TeX-master: "master"
%%% End:

\section*{Acknowledgments}

Marcel Bollmann was supported by Deutsche Forschungsgemeinschaft (DFG), Grant DI~1558/4.

\bibliographystyle{acl}
\bibliography{master}

\end{document}